\documentclass[letterpaper, 10 pt, conference]{ieeeconf}  
\IEEEoverridecommandlockouts                             
\usepackage{stfloats}
\usepackage{subcaption}
\usepackage{algorithm}
\usepackage{algorithmic}
\usepackage{amsfonts}
\usepackage{amsmath}
\usepackage{amssymb}
\usepackage{arydshln}
\usepackage{color}
\usepackage{float}
\usepackage{graphicx}
\usepackage{multirow}
\usepackage{textcomp}
\usepackage{threeparttable}
\usepackage{xcolor}
\usepackage{bm}
\usepackage{hyperref}

\usepackage{amsthm}

\newtheorem{problem}{Problem}
\newtheorem{theorem}{Theorem}
\newtheorem{remark}{Remark}
\newtheorem{assumption}{Assumption}

\newcommand{\expm}{\exp_{\rm m}}
\newcommand{\dd}{\mathop{}\!\mathrm{d}}

\usepackage{cuted} 

\title{\LARGE \bf
Supervisory Measurement-Guided Noise Covariance Estimation
}

\author{Haoying Li, Yifan Peng, Xinghan Li, Junfeng Wu
\thanks{	}
}    

\begin{document}
\maketitle
\thispagestyle{empty}
\pagestyle{empty}
\begin{abstract}
Reliable state estimation hinges on accurate specification of sensor noise covariances, which weigh heterogeneous measurements.
In practice, these covariances are difficult to identify due to environmental variability, front-end preprocessing, and other reasons.
We address this by formulating noise covariance estimation as a bilevel optimization that, from a Bayesian perspective, factorizes the joint likelihood of so-called odometry and supervisory measurements, thereby balancing information utilization with computational efficiency.
The factorization converts the nested Bayesian dependency into a chain structure, enabling efficient parallel computation: at the lower level, an invariant extended Kalman filter with state augmentation estimates trajectories, while a derivative filter computes analytical gradients in parallel for upper-level gradient updates. The upper level refines the covariance to guide the lower-level estimation.
Experiments on synthetic and real-world datasets show that our method achieves higher efficiency over existing baselines.
\end{abstract}

\section{INTRODUCTION}\label{sec:introduciton}

Robotic state estimation seeks to recover accurate states such as pose and velocity from noisy multi-sensor data.
This is often posed as an optimization problem over sensory data, typically using Maximum Likelihood Estimation~(MLE) or Maximum a Posteriori~(MAP) inference, where the noise covariance quantifies measurement uncertainty and governs the weighting of measurements, with higher covariances corresponding to less reliable data.
Therefore, precise noise covariance specification is vital for weighting measurements appropriately and mitigating unreliable information.

However, obtaining precise knowledge of noise covariance is challenging~\cite{ebadi2023present}, as sensor characteristics vary with environmental conditions and often require recalibration. Moreover, front-end preprocessing, such as feature extraction, alters raw measurements and obscures their true statistical properties~\cite{khosoussi2025joint}.
Noise covariance estimation has consequently received significant research attention.
Existing approaches span from MLE and MAP formulations that explicitly optimize noise covariances within probabilistic models~\cite{khosoussi2025joint, tsyganova2017svd}, to performance-driven methods that tune parameters by minimizing trajectory tracking error~\cite{qadri2024learning, liu2025debiasing}.
Both gradient-based methods leveraging analytical derivative solutions~\cite{sarkka2023bayesian} and derivative-free approaches that avoid explicit gradient computation~\cite{9357964, burnett2021radar, hu2017efficient} have been explored.

\begin{figure}[t!]
     \centering
     \includegraphics[width=0.72\linewidth]{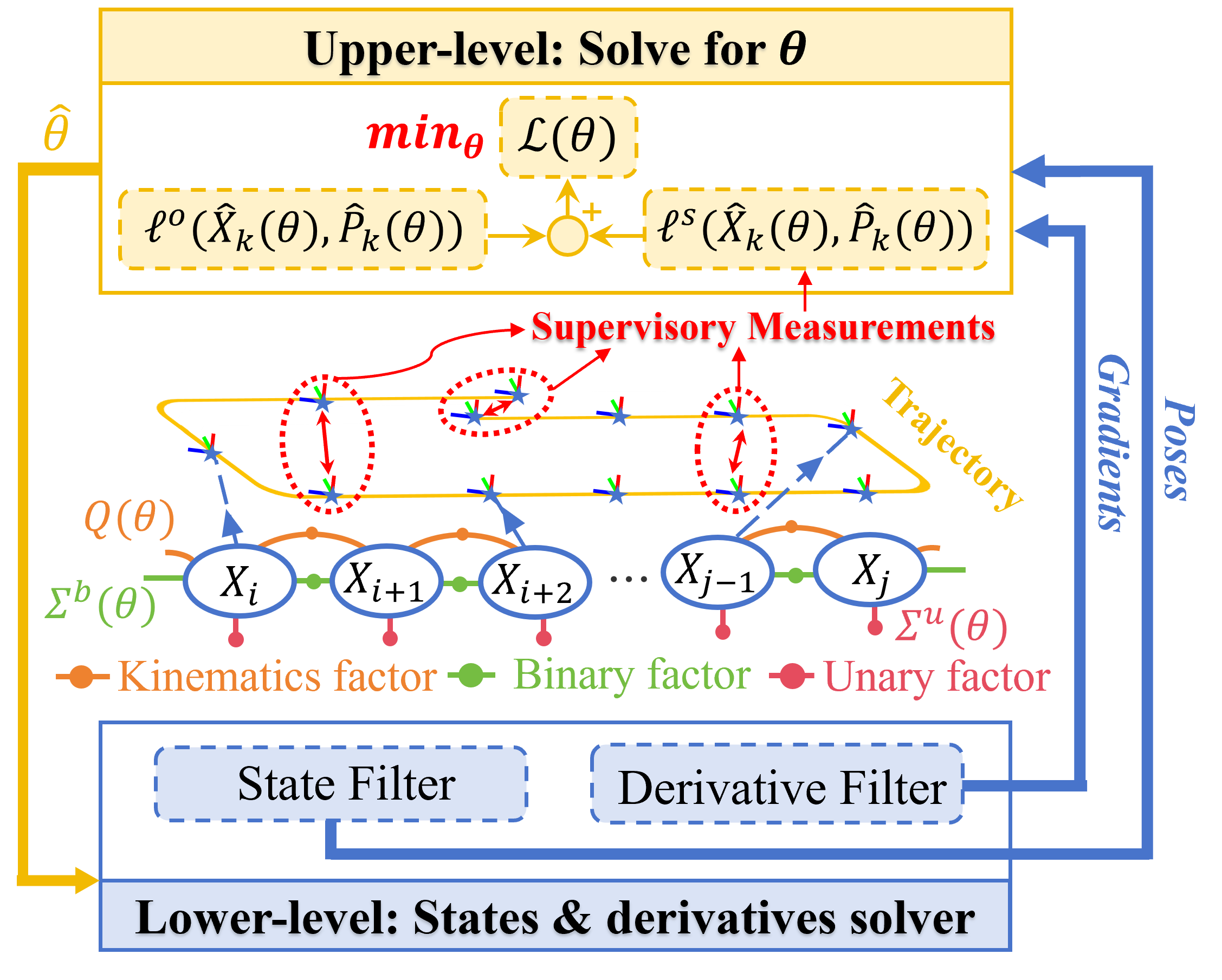}
     \caption{System framework.}
     \label{fig:framework}
 \end{figure}

Despite this progress, two challenges remain: how to fully exploit sensory information and how to estimate noise parameters efficiently. To address them, we partition measurements into odometry, for fast trajectory estimation and gradient evaluation, and supervisory, for complementary noise calibration.
Based on an ML formulation, this partition naturally leads to a bilevel optimization problem, as illustrated in Fig.~\ref{fig:framework}.
Our key contributions are:
\begin{enumerate}
    \item  \textbf{Likelihood factorization in the ML formulation.}
We distinguish the roles of odometry and supervisory measurements in forming cross-temporal correlations and separate them in the likelihood derivation.
This results in two additive terms—odometry loss and supervisory loss—where supervisory information propagates via a marginal distribution conditioned only on odometry data.
The factorization transforms the nested Bayesian network arising from loop closures in SLAM into a chain structure, enabling filter-based methods to incorporate loop closures without excessive complexity, and naturally leads to a bilevel optimization framework. 
    
\item \textbf{State Filter and Derivative Filter for bilevel updates.}  
The chain-structured Bayesian network allows a \textit{State Filter} and a \textit{Derivative Filter} to run in parallel, producing lower-level states and upper-level implicit gradients.  
Supervisory information influences the solution through covariance–triplet updates rather than nested Bayesian dependencies, reducing complexity.  
Both filters can be implemented using standard Kalman filtering with state augmentation.

    \item \textbf{Simulation and Experiment Validation}. We evaluate the proposed method against baseline approaches on synthetic and real-world datasets, demonstrating accuracy and improved efficiency.  
\end{enumerate}

\section{Related Works}
This section reviews related work on noise covariance estimation, classified by the objectives they optimize.

\textit{1) MAP/MLE methods}: Noise covariance optimization via MLE/MAP formulations has been extensively studied.
While derivative-free methods like expectation maximization jointly optimize parameters and states under likelihood objectives~\cite{9357964,burnett2021radar}, they require full posterior approximation, while point estimation is computationally preferable in practice.
Gradient-based methods for parameter estimation in Kalman filtering minimize the negative log-likelihood, also known as the energy function~\cite{sarkka2023bayesian}, using only the observations directly employed by the filter. The required gradients can be obtained either through forward sensitivity equations~\cite{tsyganova2017svd} or via backpropagation~\cite{parellier2023speeding}. 
In~\cite{khosoussi2025joint}, states and noise covariances are jointly optimized under a MAP formulation by exploiting the convexity of the joint estimation problem and using an alternating optimization scheme, with most of the computational cost incurred in solving for the states.

\textit{2) Performance-driven methods}: 
Several existing works estimate noise parameters under the supervision of state-estimation performance.
The method in~\cite{li2019unsupervised} optimizes innovation-based objectives, using measurement residuals in the outer loop and state estimation in the inner loop.
The work in~\cite{wang2023neural} proposes an adaptive disturbance estimation for a moving horizon estimator via bilevel optimization with fast derivative computation, while it requires a reference trajectory.
The framework in~\cite{qadri2024learning} formulates covariance learning as constrained bilevel optimization, which requires ground-truth supervision and uses numerical gradients.
Parameters can be optimized to fit the current task using these approaches. However, without probabilistic modeling, the resulting parameters may deviate from the true underlying values.

Extending the aforementioned works, we cast noise parameter estimation as an MLE problem, offering a principled probabilistic basis. We further partition the measurements and design distinct upper- and lower-level objectives, enhancing efficiency while leveraging richer information.


\section{Preliminary on Robotic Lie Groups}
This section presents the preliminaries on matrix Lie groups used in the derivation of this work, following~\cite{sola2018micro}.

Let $\mathbf{G}$ be a matrix Lie group with Lie algebra $\mathfrak{g}$. Note that 
$\mathfrak g$ is isomorphic to $\mathbb R^{\dim \mathfrak g}$.
The exponential map 
$
\exp: \mathbb{R}^{\dim \mathfrak g} \rightarrow \mathbf{G}, \quad \xi \mapsto \expm(\xi^\wedge)
$
relates $\mathfrak{g}$ to its associated group $\mathbf{G}$, 
where $(\cdot)^\wedge$ denotes the skew operator and $\expm(\cdot)$ the matrix exponential.
In a neighborhood of the identity, $\exp$ is locally invertible, enabling the definition of the logarithm map
$\log: \mathbf{G} \rightarrow  \mathbb{R}^{\dim \mathfrak g}, \quad X \mapsto \log(X)^\vee,
$
where $(\cdot)^\vee$ is the inverse of $(\cdot)^\wedge$.
The special orthogonal group $SO(3)$ represents the set of all possible rotations of a rigid body in three-dimensional space. 
The special Euclidean group $SE(3)$ represents rigid-body transformations, defined as
$$
SE(3)\triangleq\left\{
\left[\begin{array}{cc}
R & p\\
0 &1
\end{array}\right]\;\middle|\;
R\in SO(3), p\in\mathbb R^{3}\right\}.
$$
When velocity is further incorporated into  rigid-body motion, the resulting group is $SE_2(3)$, given by
\begin{equation}\label{def_SE2_3}
SE_2(3) \triangleq\left\{
\left[
\begin{array}{cc}
R & p ~~~ v \\ \bm 0 & I
\end{array}
\right]
\middle|
R \in SO(3), p,v \in \mathbb{R}^3
\right\},
\end{equation}
where $\bm 0$ and $I$ denote the zero matrix and the identity matrix, with the dimension omitted when clear from context.
For $x^{\wedge}, y^{\wedge} \in \mathfrak{g}$ and $\|x\|$ small, their compounded exponentials can be approximated to
$
 \exp (x) \exp (y) \approx \exp ( \operatorname{dexp}_{y}^{-1} x+y),
$
where $\operatorname{dexp}_x$ is the left Jacobian of $x$.

Let $\mathcal{M}$ be the smooth manifold of a Lie group.  
For any $X_1, X_2 \in \mathcal{M}$ and tangent vector $\xi \in \mathbb{R}^{\dim \mathcal{M}}$, we define the \emph{plus} and \emph{minus} operators, $X_1 \boxplus \xi$ and $X_1 \boxminus X_2$, which map between the manifold and its tangent space.
These operators admit left and right versions; the choice is application-dependent and left implicit. For instance, in the left case,
$
 X_1 \boxplus \xi =  \exp(\xi){X}_1, X_1  \boxminus X_2 = \log(X_1X_2^{-1}).
$
On product manifolds, the operators act componentwise~\cite{sola2018micro}.
The uncertainty of $X \in \mathcal{M}$ is modeled in the tangent space by a Gaussian distribution~\cite{barfoot2014associating}.  
Let $\xi \sim \mathcal{N}(0,\Sigma)$, the induced distribution is
\begin{align}
 \mathcal{N}_L(\bar{X}, \Sigma) \triangleq 
 \eta \exp\!\left(-\tfrac{1}{2} (X \boxminus \bar{X})^\top \Sigma^{-1} (X \boxminus \bar{X})\right),
 \label{eq:prob}
\end{align}
where $\eta$ is the normalization coefficient.


\section{Problem Statement}
In this section, we formally state the noise covariance estimation problem.

Consider a mobile robot whose kinematic evolution follows a group-affine discrete-time system as:
\begin{equation}
X_{k+1} = f(X_{k}, u_{k}, w_{k}), ~ w_{k}\sim \mathcal{N}(\bm0,Q(\theta)),\label{eq:process}
\end{equation}
where $X_k \in \mathcal{X}$ denotes the robot's state at time step $k$ on the Lie group $\mathcal{X}$, 
$u_k \in \mathcal{U}$ is the control input, 
and $w_k$ is the process noise.
Conventionally, introspective sensor measurements (e.g., from IMUs) are used as inputs to the motion model through preintegration, which conflates uncertainty in measurements with the process noise $w_k$~\cite{burnett2024imu}. 

The motion of a robot can be captured using extrospective sensors. In this paper, we categorize their measurements into two types~\cite{dellaert2017factor}.
Unary measurements, including those from GPS receivers, correspond to a single motion state and provide information in a fixed reference frame, whereas binary measurements, including LiDAR or RGB-D point clouds from iterative closest point~(ICP), relate two consecutive motion states.
At this stage, we do not constrain the sensor types to specific implementations to maintain theoretical completeness. The probabilistic models for these two measurement classes are 
\begin{align}
y^b_{k}=  &h(X_{k-1}, X_{k}) \boxplus \nu^{b}_k,~&&\nu^{b} \sim \mathcal N(\bm 0, \Sigma^{b}(\theta)),\label{eq:binary}\\
y^u_k  =  &g(X_{k}) \boxplus  \nu^{u}_k,~ &&\nu^{u} \sim \mathcal N(\bm 0, \Sigma^{u}(\theta)), \label{eq:unary}
\end{align}
where $y_k^u \in \mathcal{Y}^u, y_k^b\in \mathcal{Y}^b$ with $\mathcal{Y}^u$ and $\mathcal{Y}^b$ denoting the respective output spaces.  
Let $\bm{y}^o_{1:k} = \{y_i^u, y_i^b\}_{i=1:k}$ denote the set of unary and consecutive binary observations up to time $k$, which are collectively referred to as \emph{odometry measurements}, since they only provide local motion information and are susceptible to drift accumulation over time.

As discussed in Section~\ref{sec:introduciton}, accurate knowledge of process and measurement noise covariances is crucial for odometry but remains difficult to determine in practice.
Consequently, we treat these covariances as unknown quantities and introduce a parameterization via a variable $\theta$. 
This parameterization may take different forms, from simple vectorization of matrix entries (with symmetry, positive definiteness, and other constraints enforced) to more expressive representations based on neural networks.
While some prior works~\cite{qadri2024learning,liu2025debiasing} leverage ground-truth robot poses as supervisory signals to learn the noise parameters, in this paper, we propose a general approach. We introduce a new class of measurements—termed \emph{supervisory measurements}—that directly inform the robot's relative poses over extended trajectories. 
These measurements provide supervision for estimating $\theta$. 
Formally, the supervisory measurement model is
\begin{equation}~\label{eq:superO}
y_{ij}^s = s(X_i,X_j)\boxplus \nu_{ij},\quad \nu_{ij}\sim \mathcal{N}(\bm 0,\psi),
\end{equation}
where $\psi$ denotes the covariance and is assumed to be known. 
Typical examples include loop closures and ground-truth poses, the latter can be interpreted as $y_{0j}^s$, i.e., relative poses with respect to the initial pose or another fixed frame.

\begin{remark}
Ground-truth poses, as supervisory measurements obtained from motion capture systems or other high-precision sources, are typically modeled with negligible or zero uncertainty.
Under this assumption, one can set $\psi=\bm 0$ in~\eqref{eq:superO}.
This corresponds to a Dirac delta probability, which arises as the limit of a sequence of Gaussian distributions centered at the origin with variances tending to zero. \qed
\end{remark}

Define the state projection
$
\pi(y_{ij}^s)=\{X_i,X_j\},
$
which takes a supervisory measurement 
$y_{ij}^s$ and returns the set of 
states invovled in $y_{ij}^s$.
Similarly, the index map is defined as 
$\pi_{\rm id}(y_{ij}^s)=\{i,j\}$.
Letting
$\mathcal{I}^s_{k}=\{i\in 1,\ldots,k\mid \{i,j\}= \pi_{\rm id}(y_{ij}^s), \exists j\in 1,\dots,k\}$,
 then $\bm{y}^s_{1:k}\triangleq\{y_{ij}\}_{\{i,j\}\in \mathcal I^s_k}$.

Given odometry measurements $\bm{y}^o_{1:N}$ and supervisory measurements $\bm{y}^s_{1:N}$ collected up to time~$N$, the noise covariance estimation problem can be formulated as a MLE problem
\begin{equation}\label{eqn:MLE_problem}
\hat{{\theta}} = \operatorname*{argmax}_{{\theta}\in\Theta} p(\bm{y}^o_{1:N}, \bm{y}^s_{1:N} \mid \theta),
\end{equation}
 where $\hat{\theta}$ is termed the MLE estimate to~$\theta$ and $\Theta$ represents the admissible parameter space. 

To ensure clarity in subsequent discussions, some notational conventions are clarified below.
The set of states up to time $N$ is denoted in bold italic as \(\bm{X}_{1:N} = \{X_1, \ldots, X_N\}\). Within this set, \(\bm{X}^s_{1:N}\) represents the subset of states involved in supervisory measurements, defined as
\[
\bm{X}^s_{1:N} = \{X_i \mid X_i \in \pi(z_{ij}), \exists z_{ij} \in \bm{y}^s_{1:N}\}.
\]
For notational simplicity, we may omit the time subscript $1:N$ or $N$ from the above symbols whenever the time horizon is clear from the context.
The remaining states, excluding those in \(\bm{X}^s\), are denoted as \(\bm{X}^o= \bm{X} \setminus \bm{X}^s\).


\section{Methodology}

Our subsequent derivations are based on the following assumption regarding the measurement and process noises in the system. 
\begin{assumption}~\label{ass:indep}
The noises $w_k$'s, $\nu^b_k$'s, $\nu^v_k$'s, and $\nu_{ij}$'s are mutually independent random variables.\qed
\end{assumption}
The assumption is standard in SLAM literature~\cite{khosoussi2025joint,zheng2024fast}, which enables tractable theoretical analysis while remaining practically justifiable for most real-world applications.

\subsection{Problem Formulation}
By introducing the motion states as latent variables and invoking Assumption~\ref{ass:indep}, the likelihood~\eqref{eqn:MLE_problem} is factorized as
\begin{align*}
&p(\bm{y}^o,\bm{y}^s \mid \theta) \\
=& \int \int p(\bm{y}^o,\bm{y}^s,\bm{X}^o,\bm{X}^s \mid \theta)\, \mathrm{d} \bm{X}^o \mathrm{d} \bm{X}^s \\
= &\int \int 
p(\bm{y}^s \mid \bm{X}^o, \bm{X}^s, \bm{y}^o, \theta)\, p(\bm{X}^o, \bm{X}^s, \bm{y}^o \mid \theta)
\, \mathrm{d} \bm{X}^o \mathrm{d} \bm{X}^s \\
=& \int \int 
p(\bm{y}^s \mid \bm{X}^s)\, p(\bm{X}^o, \bm{X}^s \mid \bm{y}^o, \theta)\, p(\bm{y}^o \mid \theta)
\, \mathrm{d} \bm{X}^o \mathrm{d} \bm{X}^s,
\end{align*}
where the integrations are taken over some product spaces of 
$\mathcal X$ by default. 
Further marginalizing out $\bm{X}^o$ yields
\begin{align*}
&p(\bm{y}^o,\bm{y}^s \mid \theta) \\
=& p(\bm{y}^o \mid \theta) \int  
p(\bm{y}^s \mid \bm{X}^s)\,\left(\int p(\bm{X}^o, \bm{X}^s \mid \bm{y}^o,\theta)\,
\mathrm{d} \bm{X}^o \right)\mathrm{d} \bm{X}^s \\
=& p(\bm{y}^o \mid \theta) \int 
p(\bm{y}^s \mid \bm{X}^s)\, p(\bm{X}^s \mid \bm{y}^o, \theta)\,
\mathrm{d} \bm{X}^s.
\end{align*}
The negative log-likelihood is decomposed into two terms
\begin{align*}
{{L}}(\theta)\triangleq-\log p(\bm{y}^o,\bm{y}^s \mid \theta) 
= \ell^o(\theta) + \ell^s(\theta),
\end{align*}
where $ \ell^o(\theta) \triangleq -\log p(\bm{y}^o \mid \theta)$ and
\begin{equation}
     \ell^s(\theta) \triangleq -\log \int p(\bm{y}^s \mid \bm{X}^s)\, p(\bm{X}^s \mid \bm{y}^o,\theta)\,\dd \bm{X}^s\label{eq:ls}.
\end{equation}
We refer to $\ell^o(\theta)$ as the \emph{odometry loss} and $\ell^s(\theta)$ as the \emph{supervisory loss}, with $\ell^o(\theta)$ admits the following factorization
\begin{equation}\label{eq:lo1}
    \ell^o(\theta)
=   -\sum_{k=1}^N \log p\left(y_k^o \mid \bm{y}^o_{1: k-1}, \theta\right),
\end{equation}
where it further holds that
\begin{align*}
p\left({y}_k^o \mid \bm{y}_{1:k-1}^o, \theta\right) = \int p\left({y}_k^o \mid X_k, \theta\right) p\left(X_k\mid \bm{y}^o_{1:k-1}, {\theta}\right) \mathrm{d} X_k.
\end{align*}
where $p({y}_k^o \mid X_k, \theta)$ represents the measurement models~\eqref{eq:binary}~\eqref{eq:unary}, and $p\left(X_k \mid \bm{y}_{1:k-1}^o, \theta\right)$ is the Gaussian predictive distribution.
With the above factorization in place, the distribution $p\left(X_k \mid \bm{y}_{1:k-1}^o, \theta\right)$ and $p(\bm{X}^s \mid \bm{y}^o,\theta)$ remain to be derived.
We tone down the hope of 
acquiring exact computation for them, for exact analytical expressions for these distributions are generally intractable and rarely available in closed form.  Therefore, in what follows, we compute them by resorting to the local linearization of the measurement model within the tangent space of $X_k$.

\subsection{State Filter}\label{sec:SF}
This part presents the design of an invariant extended Kalman filter~(InEKF) that recursively computes $p\left(X_k \mid \bm{y}_{1:k-1}^o, \theta\right)$ with states augmentation. As will be shown in the subsequent derivation, the augmented state in the InEKF enables the calculation of $p(\bm{X}^s \mid \bm{y}^o,\theta)$.

To keep track of the correlation between states in $\bm{X}_k^s$, first define the following composite state
$$
Y_k=[X_{k-1}^\top, X_{k}^\top,
X_{i_{k,1}}^\top,\ldots, X_{i_{k,l}}^\top
]^\top.
$$
In the above definition, $X_{i_{k,1}},\ldots, X_{i_{k,l}}$ collectively constitute $\bm{X}^s_k$, 
with a predetermined index set $\mathcal{I}^s_k$ (e.g., the indices of keyframe candidates selected for loop closures).

The notations used in the state filter are clarified here.
Let $\bar{Y}_k,\bar{P}_k$ and $\hat{Y}_k,\hat{P}_k$ denote the prior and the posterior estimates with the estimation error covariance of $Y_k$. 
The estimates and associated errors are given by
\begin{align*}
    \bar{Y}_k =&[\hat{X}_{k-1|k-1}^\top, \bar X_{k}^\top,\bar X_{i_{k,1}}^\top,\ldots, \bar X_{i_{k,l}}^\top ]^\top,~\bar\zeta_k \triangleq Y_k\boxminus \bar{Y}_k,\\
    \hat{Y}_k =& [\hat{X}_{k-1|k}^\top, ~\hat{X}_{k|k}^\top,\hat X_{i_{k,1}}^\top,\ldots, \hat X_{i_{k,l}}^\top ]^\top,~\hat\zeta_k \triangleq Y_k\boxminus \hat{Y}_k,
\end{align*}
where the state estimate at time $k-1$ after incorporating $y^b_{k-1}$ is denoted by $\hat{X}_{k-1|k-1}$, 
while $\hat{X}_{k-1|k}$ refers to the estimate obtained after incorporating $y^b_{k}$.

Upon receiving the input $u_{k-1}$, while other states remain unchanged, $\hat{X}_{k-1|k-1}$ is propagated through the nominal part of~\eqref{eq:process} as follows:
\begin{equation}\label{eq:propX}
    \bar{X}_k = f(\hat{X}_{k-1|k-1}, u_{k-1}, \bm 0)
\end{equation}
The resulting a priori estimation error and its covariance can be written in partitioned form as
\begin{equation*}
\bar\zeta_k =[(\bar{\zeta}_k^o)^\top~~(\bar{\zeta}_k^s)^\top ]^\top,\quad \bar P_k =\begin{bmatrix}
   \bar P^o_k & \bar P_k^{os}\\
   \bar P_k^{so} &\bar P_k^s
\end{bmatrix},
\end{equation*}
where $(\cdot)^o$ corresponds to the two-frame odometry states at time step $k$, i.e., $X_{k-1}, X_{k}$, and $(\cdot)^s$ relates to the $\bm{X}_k^s$. 
The estimation error covariance matrix evolves as
\begin{equation}\label{eq:prop_P}
\bar{P}_k = F_{k-1} \hat{P}_{k-1}  F_{k-1}^\top + Q_{0,k},
\end{equation}
where $F_{k-1} = \bigl[\begin{smallmatrix}
    \Phi_{k-1} & \bm 0 \\ \bm 0 & I
\end{smallmatrix}\bigr]$, 
$Q_{0,k} = \bigl[\begin{smallmatrix}
    Q_0 & \bm 0 \\\bm 0 &\bm 0
\end{smallmatrix}\bigr]$, 
with $\Phi_{k-1} \triangleq \bigl[\begin{smallmatrix}
 \bm   0 & I \\ \bm   0 & \phi_{k-1}
\end{smallmatrix}\bigr]$, $Q_0 \triangleq \bigl[\begin{smallmatrix}
   \bm   0 & \bm   0\\ \bm   0 & Q
\end{smallmatrix}\bigr]$.
Note that
$\phi_{k-1}$ follows~\cite{barrau2016invariant} and owing to the log-linear property of group-affine dynamics, $\phi_{k-1}$ is independent of ${X}_{k-1}$.

The update step serves to incorporate odometry measurements.
For generality, unary and binary measurements together form a stacked residual
\begin{equation}\label{eq:rk}
    r_k = [(r^u_k)^\top~~(r^b_k)^\top]^\top,
\end{equation}
where $r^u_k \triangleq y_k^u \boxminus g(\bar{X}_k)$ and $r^b_k \triangleq y^b_k \boxminus h(\hat X_{k-1|k-1},\bar X_{k}) $. In practice, only the available measurements are included in this step.
The residual Jacobian and the corresponding measurement noise covariance take the form
\begin{align}
H_k =
\begin{bmatrix}
H^u_k \\
H^b_k
\end{bmatrix}, \Sigma_k =
\begin{bmatrix}
\Sigma^u & 0 \\
0 & \Sigma^b
\end{bmatrix},
\end{align}
where
$H^b_k \triangleq \begin{bmatrix}
   \left. \tfrac{\partial r^b_k(\hat{X}_{k-1|k-1} \boxplus \xi)}{\partial \xi}\right|_{\xi=\bm 0} &    \left. \tfrac{\partial r^b_k(\bar{X}_{k} \boxplus \xi)}{\partial \xi}\right|_{\xi=\bm 0}
\end{bmatrix}
$ and  $
H^u_k \triangleq 
\left.\tfrac{\partial r^u_k(\bar{X}_k \boxplus \xi)}{\partial \xi}\right|_{\xi=\bm0}$. 
Let ${H}_{0,k} = [H_k ~~\bm 0]$, it follows that
\begin{align}
S_k =& {H}_{0,k} \bar{P}_k {H}_{0,k}^\top + \Sigma_k = H_k \bar{P}^o_k H_k^\top +\Sigma_k,\label{eq:S}\\
K_k =&\bar{P}_k H_{0,k}^\top S_k^{-1},~~\check{P}_k = \bar{P}_k-K_k H_{0,k} \bar{P}_k\label{eq:K}\\
\hat{Y}_k =& \bar{Y}_k\boxplus (K_kr_k).\label{eq:updateY}
\end{align}

Once the update completes, $\hat{X}_k$ will append to the end of $\hat{Y}_k$ if $X_k \in \bm{X}_k^s$, resulting in the covariance update
\begin{equation}\label{eq:copyP}
    \hat{P}_k=J_k \check{P}_k  J_k^\top,
\end{equation}
where $J_k = [I~ J^\top]^\top$
with {$J = [I \; \bm 0]$ } when the appending takes place, and 
$J_k = I$, otherwise.

\begin{remark}[Dimension Control.]
   In the absence of unary measurements, cross-temporal correlations vanish and state augmentation can be omitted; otherwise, the number of augmented states can be deliberately limited to control computational complexity. In practice, calibration trajectories are typically short, and downsampling can be applied if needed. Specifically, with $M$ states for supervisory measurements, yields at most $M(M-1)/2$ pairwise distinct observations---generally sufficient to support reliable calibration.\hfill $\square$
\end{remark}

\subsection{\texorpdfstring{Derivatives of $\ell^{o}$ and $\ell^s$ w.r.t.\ $\theta$}{Derivatives of lo and ls w.r.t. theta}}

By substituting the Gaussian predictive distribution from Subsection~\ref{sec:SF} into~\eqref{eq:lo1} and discarding additive constants, the odometry loss takes the closed-form expression~\cite{sarkka2023bayesian}
\begin{equation}\label{eq:loo}
    \ell^o(\theta) \cong \sum_{k=1}^N \frac{1}{2} \log| S_k(\theta)|+\frac{1}{2}r_k(\theta)^\top (S_k(\theta))^{-1}r_k(\theta),
\end{equation}
where $r_k$ is from~\eqref{eq:rk} and $S_k$ is from~\eqref{eq:S} and the symbol $\cong$ denotes equivalence up to the omitted additive constants.

For ease of presentation, we define
\[
l_k^o(\theta) \triangleq \tfrac{1}{2}\log |S_k(\theta)| 
+ \tfrac{1}{2} r_k(\theta)^\top S_k(\theta)^{-1} r_k(\theta).
\]
Evaluating the gradient of $\ell^o(\theta)$ with respect to $\theta$ requires 
the derivatives of $S_k$ and $r_k$,
with respect to $\theta$, 
which in turn necessitates a sensitivity analysis of the state filter. 
Following~\cite{sarkka2023bayesian}, applying a term-wise differential to the state filter~\eqref{eq:prop_P}–\eqref{eq:copyP}, which we refer to as the \textit{derivative filter}.

\begin{figure*}[!b] 
\noindent\rule{\textwidth}{0.4pt}\par
\vspace{-0.5em}
\setlength{\jot}{0.5pt} 
\noindent\small
\begin{minipage}{0.65\textwidth}
\begin{align}
\frac{\partial \bar{P}_k}{\partial \theta_j} &= F_k \frac{\partial\hat{P}_{k-1}}{\partial \theta_j} F_k^\top 
+ \frac{\partial {Q}_{0,k}}{\partial \theta_j}, 
\quad\frac{\partial  H_k}{\partial \theta_j} 
= \frac{\partial  H_k}{\partial \bar{\zeta}_k} 
  \frac{\partial \bar{\zeta}_k}{\partial {\theta}_j}, \notag \\
\frac{\partial S_k}{\partial\theta_j} &=
{H}_k \frac{\partial \bar{P}_k}{\partial \theta_j} {H}_k^\top 
+ \left( \frac{\partial {H}_k}{\partial \theta_j} \right) \bar{P}_k {H}_k^\top
+ {H}_k \bar{P}_k \left( \frac{\partial {H}_k}{\partial \theta_j} \right)^\top 
+ \frac{\partial \Sigma_k}{\partial \theta_j}, \label{eq:dSk}\\
\frac{\partial W_k}{\partial \theta_j} &= - W_k \frac{\partial S_k}{\partial\theta_j}W_k,
 \quad \frac{\partial K_k}{\partial \theta_j} 
= \frac{\partial \bar{P}_k}{\partial \theta_j} {H}_k^\top W_k
+ \bar{P}_k \left( \frac{\partial {H}_k}{\partial \theta_j} \right)^\top W_k
+ \bar{P}_k {H}_k^\top \frac{\partial W_k}{\partial \theta_j},\notag  \\
\frac{\partial \check{P}_k}{\partial \theta_j} 
&= \frac{\partial \bar{P}_k}{\partial \theta_j}
- \frac{\partial K_k}{\partial \theta_j} {H}_k \bar{P}_k
- K_k \left( \frac{\partial {H}_k}{\partial \theta_j} \right) \bar{P}_k
- K_k {H}_k \frac{\partial \bar{P}_k}{\partial \theta_j},\label{eq:dhatOmega}\\
\frac{\partial \hat{P}_k}{\partial \theta_j} 
&= J_k \frac{\partial  \check{P}_k}{\partial \theta_j} J_k^\top. \notag
\end{align}
\end{minipage}\hfill
\begin{minipage}{0.33\textwidth}
\begin{align}
\frac{\partial \bar{\zeta}_k}{\partial \theta_j} &= F_k \, \frac{\partial \hat{\zeta}_{k-1}}{\partial \theta_j},\notag\\
\frac{\partial{r_k}}{\partial \theta_j} &= -{H}_k \frac{\partial \bar{\zeta}_{k}}{\partial \theta_j},\label{eq:drk} \\
\frac{\partial \hat{\zeta}_k}{\partial \theta_j} &= \mathcal{J}_k\frac{\partial \bar{\zeta}_k}{\partial \theta_j} 
+ \frac{\partial K_k}{\partial \theta_j} r_k+K_k\frac{\partial r_k}{\partial \theta_j},\notag\\
\frac{\partial \hat{\zeta}_k}{\partial \theta_j} 
&= J_k \frac{\partial \hat{\zeta}_k}{\partial \theta_j}, \notag
\end{align}
where $\tfrac{\partial\hat P_0}{\partial \theta_j} = 0$ and $\tfrac{\partial\hat \zeta_0}{\partial \theta_j} = 0$ at initialization, and $\mathcal{J}_k\triangleq \operatorname{dexp}_{(K_kr_k)}^{-1}$.
\end{minipage}
\end{figure*}

\begin{theorem}[Derivative of $\ell^o$]\label{Theo:lo_dlo}
The derivative of  $\ell^o({\theta})$ with respect to each component $\theta_j$ of $\theta$ is the sum of its derivatives from each time step:
\begin{equation}\label{eq:dlo}
 \frac{\partial \ell^o({\theta})}{\partial \theta_j}
 =\sum_{k=1}^N \frac{\partial l_k^o({\theta})}{\partial \theta_j}.
\end{equation}
Each constituent term $\frac{\partial l_k^o({\theta})}{\partial \theta_j}$ has the form
\begin{equation}\label{eq:dvarphi}
\frac{1}{2}\,\mathrm{tr}( S_k^{-1} \frac{\partial S_k}{\partial \theta_j} ) 
+ ( \frac{\partial r_k}{\partial \theta_j} )^\top S_k^{-1} r_k - \frac{1}{2}\, r_k^\top S_k^{-1} ( \frac{\partial S_k}{\partial \theta_j} ) S_k^{-1} r_k,
\end{equation}
with $S_k({\theta})$ and $r_k({\theta})$ abbreviated as $S_k$ and $r_k$. The computation of  $\partial S_k / \partial \theta_j$ and $\partial r_k / \partial \theta_j$ follows~\eqref{eq:dSk} and~\eqref{eq:drk}.$\square$
\end{theorem}

The following describes the computation of $\ell^s$ and the derivation of its gradient. Specifically, the MAP estimate from $p(\bm{X}^s \mid \bm{y}^o,\theta)$ is $[\hat{X}{i_{N,1}}^\top, \ldots, \hat{X}{i_{N,l}}^\top]^\top$ with $\hat{P}_N^s$ extracted from the output of the state filter. For notational simplicity, we define $\hat{\bm X}^s\triangleq[(\hat{X}_1^s)^\top, \ldots, (\hat{X}_l^s)^\top]^\top$. 


For each $y_{ik}^s \in \bm y^s$, the residual is defined as
$v_{ik} = y_{ik}^s \boxminus s(\hat X_i^s,\hat X_k^s)$, 
and its Jacobian w.r.t.\ the error state of $\hat Y^s$ has the following form 
\begin{equation}
  H_{ik}^s = \left[\bm 0~\tfrac{\partial v_{ik}(\hat X_i^s \boxplus \xi)}{\partial \xi}\big|_{\xi=\bm 0}~\bm0~\tfrac{\partial v_{ik}(\hat X_k^s \boxplus \xi)}{\partial \xi}\big|_{\xi=\bm 0}~\bm0\right]  
\end{equation}
Stacking all residuals gives the aggregate vector and Jacobian
\begin{align}\label{eq:vs}
    v \triangleq \operatorname{col}\{v_{ik}\},
    \qquad 
    H^s \triangleq  \operatorname{col}\{H_{ik}^s\},
\end{align}
where $\operatorname{col}(\cdot)$ stacks vectors or block rows vertically.

Given the above stacked measurements associated with the supervisory states, the uncertainty of $\bm{X}^s$ is modeled in the tangent space as in~\eqref{eq:prob}, given by
$ p(\bm{y}^s \mid \bm{X}^s)= \mathcal{N}(\bm{y}^s \mid H^s \zeta^s_N, \Psi)$ and $p(\bm{X}^s \mid \bm{y}^o, \theta) = \mathcal{N}(\zeta^s_N \mid \hat{\zeta}^s_N, \hat{P}^s_N)$.
Substituting these distributions into~\eqref{eq:ls} yields the expression for $\ell^s(\theta)$. By applying the chain rule, Theorem~\ref{Theo:ls_dls}  establishes the derivative of $\ell^s$ with respect to $\theta$.

\begin{theorem}[Derivative of $\ell^s$] \label{Theo:ls_dls}
The supervisory loss $\ell^s$ is computed as
\begin{equation}\label{eq:lss}
    \ell^s(\theta) \cong \frac{1}{2}\log |C(\theta)| 
    + \frac{1}{2} v(\theta)^\top C(\theta)^{-1} v(\theta),
\end{equation}
where $v$ is given by~\eqref{eq:vs} and  
\(C = H^s \hat{P}_N^s(\theta)(H^s)^\top + \Psi\)
with $\Psi$ denoting the block-diagonal matrix formed from $\psi$.
The derivative of $\ell^s$ with respect to~$\theta_j$ is 
\begin{align}
    \frac{\partial \ell^s(\theta)}{\partial \theta_j} 
    =&\;\frac{1}{2}\operatorname{tr}\!\left(C^{-1} \frac{\partial C}{\partial \theta_j}\right)
       + \left(\frac{\partial v}{\partial \theta_j}\right)^\top C^{-1} v \notag\\
      &\; - \frac{1}{2} v^\top C^{-1} \left(\frac{\partial C}{\partial \theta_j}\right) C^{-1} v. \label{eq:dls}
\end{align}
where $\partial C /\partial \theta_j$ and $\partial v/\partial \theta_j$ are computed as
\begin{align*}
\frac{\partial C}{\partial \theta_j} 
=&\frac{\partial H^s}{\partial \theta_j}\hat{P}_N^s (H^s)^\top 
   + H^s \frac{\partial \hat{P}_N^s}{\partial \theta_j} (H^s)^\top 
   + H^s \hat{P}_N^s \Big(\frac{\partial H^s}{\partial \theta_j}\Big)^\top, \\
\frac{\partial v}{\partial \theta_j} 
=&- H^s \frac{\partial \hat{\zeta}^s_N}{\partial \theta_j},\quad 
\frac{\partial H^s}{\partial \theta_j} 
= \frac{\partial H^s}{\partial \hat \zeta^s_N}\, \frac{\partial \hat\zeta^s_N}{\partial \theta_j}. ~~~~~~~~~~~~~~~~~~~~ \square
\end{align*}
\end{theorem}

Combining $\ell^o(\theta)$~\eqref{eq:loo} and $\ell^s(\theta)$~\eqref{eq:lss}, the full negative log-likelihood up to an additive constant is expressed as
\begin{align*}
{L}(\theta)& \triangleq \mathcal{L}(\bar{\bm{X}}(\theta),\bar{\bm{P}}(\theta),\hat{{\bm X}}^s(\theta), \hat{{ P}}^s_N(\theta))
  \\&\cong{\ell^o}(\bar{\bm{X}}(\theta),\bar{\bm{P}}(\theta))+\ell^s(\hat{{\bm X}}^s(\theta), \hat{{ P}}^s_N(\theta)). 
\end{align*}

Consequently, the noise covariance estimation admits the following bilevel optimization formulation \footnote{To simplify notation, we allow an ambiguity by using 
$\hat{\bm{X}}(\theta)$ and $\hat{\bm{P}}(\theta)$ to denote both 
$\bar{\bm{X}}(\theta), \hat{\bm{X}}^s(\theta)$ and 
$\bar{\bm{P}}(\theta), \hat{\bm{P}}^s_N(\theta)$.}:
\begin{problem}\label{Problem:finalBO}
\begin{align*}
&\text{\small (Upper level)} ~~
\hat{\theta} = \arg\min_{\theta \in \Theta}
\mathcal{L}(\hat{\bm{X}}(\theta), \hat{\bm{P}}(\theta)), \\[0.3em]
&\text{\small (Lower level)} ~~ \text{s.t.}
\{\hat{\bm{X}}(\theta), \hat{\bm{P}}(\theta)\} =
\operatorname*{argmax}\limits_{X_k,\, 1 \leq k \leq N}  
p(X_k \mid \bm{y}^o_{1:k}, \theta) .
\end{align*}
\end{problem}

\begin{figure}[H]
    \centering
    \includegraphics[width=0.70\linewidth]{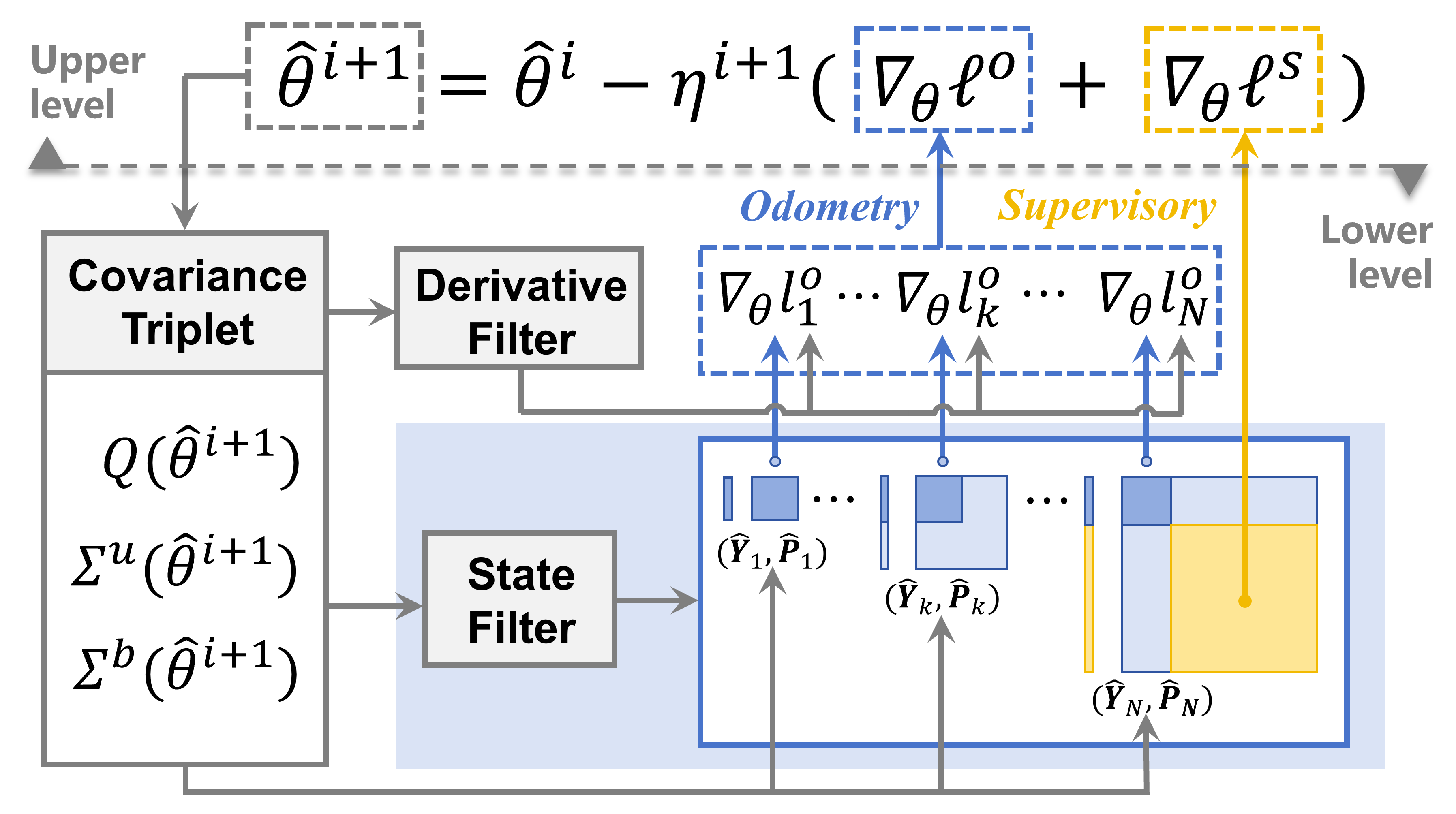}
    \caption{Logical flow for solving the bilevel problem.}
    \label{fig:method}
\end{figure}

To address Problem~\ref{Problem:finalBO}, the lower-level is solved using the state filter introduced in Section~\ref{sec:SF}.
For the upper-level problem, a gradient descent method is employed, where the gradient is given by
\begin{equation}\label{eq:dL}
    \nabla_{\theta} \mathcal{L} 
    = \nabla_{\theta}\ell^o + \nabla_{\theta}\ell^s,
\end{equation}
with $\nabla_{\theta}\ell^o$ from Theorem~\ref{Theo:lo_dlo} and $\nabla_{\theta}\ell^s$ from Theorem~\ref{Theo:ls_dls}.
In summary, the computation diagram is shown in Figure.\ref{fig:method}.

\vspace{-1em}


\begin{algorithm}[H]
\caption{Supervisory Measurement-Guided Noise Covariance Estimation}
\label{alg:plato}
\begin{algorithmic}[1]
\REQUIRE Initial state $X_0$, observations $\bm{y}^o$, $\bm{y}^s$, initial hyperparameters $\theta^{(0)}$, threshold $\epsilon$, max\_iter 
\ENSURE Noise covariance parameters $\theta$
\STATE Initialize iteration counter $i \gets 0$
\WHILE{$\|\nabla_{\theta}\mathcal{L}(\theta)\| >\epsilon$  \textbf{and} $i <$ max\_iter}
    \STATE \textbf{Lower-Level :}
        \STATE Run state filter~\eqref{eq:prop_P}-\eqref{eq:copyP} to obtain $\bm{\hat X}(\theta^{(i)})$, $\bm{\hat P}(\theta^{(i)})$, and concurrently obtain $\nabla_{\theta} \ell^o(\theta^{(i)})$ via~\eqref{eq:dlo}
    \STATE \textbf{Upper-Level :}
        \STATE Compute $\nabla_{\theta} \ell^s(\hat\theta)$ via~\eqref{eq:dls}
        \STATE Evaluate $\nabla_{\theta} \mathcal{L}(\theta^{(i)})$ using~\eqref{eq:dL}
        \STATE Update $\tilde\theta^{(i+1)} \gets \theta^{(i)} - \eta^{(i)} \,\nabla_{\theta} \mathcal{L}(\hat{\theta})$
        \STATE $\theta^{(i+1)} \leftarrow$  $\mathrm{Projection}_\Theta(\tilde\theta^{(i+1)})$
    \STATE $i \gets i + 1$
\ENDWHILE
\end{algorithmic}
\end{algorithm}

\subsection{Methodology Summary and More Discussions}

The overall execution of the noise–covariance estimation procedure is summarized in Algorithm~\ref{alg:plato}.
At the upper level, the problem is posed as a constrained optimization, where the operation $\mathrm{Projection}_\Theta$ projects the estimated parameters onto the feasible set $\Theta$.
To improve numerical stability, feasibility constraints can incorporate bounds on the eigenvalues or the condition number of the covariance matrix~\cite{qadri2024learning}.
In this work, the step size 
$\eta^{(i)}$
  is obtained using the Armijo line search rule.
More generally, once the objective function and its gradient are available, a wide range of gradient-based optimization algorithms can be employed.

Multiple parameterizations are possible for noise covariance matrices.
A widely adopted approach is Cholesky decomposition~\cite{parellier2023speeding}, which inherently guarantees positive definiteness, while the diagonal form~\cite{qadri2024learning} is a special case.
Our framework is agnostic to the specific choice of parameterization and can accommodate more expressive representations, including differentiable neural network layers~\cite{wang2023neural}. 

The proposed method is not limited to the InEKF.
For a general EKF, the required derivatives can be obtained via sensitivity equations.
We focus on the InEKF and affine group systems in this work, as they are prevalent in practice and possess favorable linearization properties~\cite{barrau2016invariant} that simplify the derivative–filter design.

Finally, we briefly relate our approach to prior work. Noise covariance estimation is often posed as alternating optimization over states and covariances, using either a shared or separate objective at two levels. We start with the covariance-level objective $L(\theta)$, which is the log-likelihood of the parameters of interest conditioned on all measurements, as derived from a Bayesian framework.
Serval prior works~\cite{qadri2024learning, liu2025debiasing, wang2023neural} tune hyperparameters, including noise covariances, by minimizing state-estimation losses against ground-truth poses, $L(\hat{\bm X}(\theta))$. Although effective in reducing trajectory error, the resulting covariances may not reflect the true noise statistics well.
To remove reliance on ground truth, \cite{hu2017introspective} introduced the approximate posterior error, which in the EKF reduces to the trace of the estimation error covariance, i.e., $L(\hat{\bm P}(\theta))$.
Innovation-based methods learn covariances from measurement residuals, leading to $L(\hat{\bm X}(\theta))$.
The traditional energy minimization approach~\cite{tsyganova2017svd,sarkka2023bayesian} corresponds to the loss $\ell^o(\theta)$, whereas our formulation incorporates the prevalent supervisory measurements in SLAM as an additional source of information to reinforce estimation performance.
More recently, \cite{khosoussi2025joint} proposed joint MAP estimation of states and noise covariances with closed-form covariance updates, but this framework is incompatible with filter-based state estimation and entails substantial cost due to repeated full-state optimization refinement.

\begin{figure*}[ht]
    \centering
    \includegraphics[width=1\linewidth]{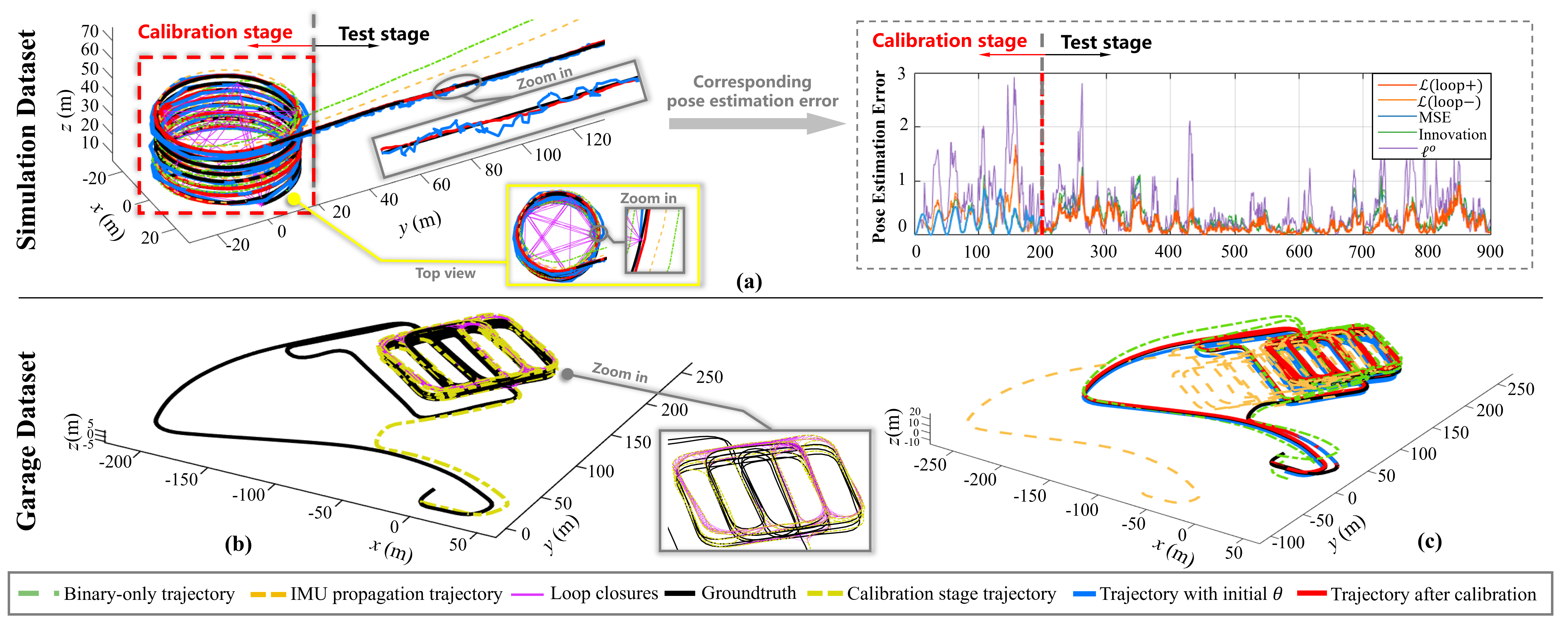}
    \caption{(a) Loop-to-Open dataset~(Synthetic). Left: trajectory with calibration results; Right: pose estimation error over time. The lowest-MSE methods are highlighted during two stages.
    (b) Visualization of the Garage dataset. (c) Trajectory comparison using only IMU or binary measurements, and trajectories before and after parameter tuning.}
    \label{fig:Simulation}
\end{figure*}

\section{Simulation and Experiment}
\subsection{Baselines}
We compare our method against three baselines.
In general, all baseline approaches adopt a bilevel structure for parameter tuning.
The lower-level problem is the same for all methods: optimizing the states conditioned on a given noise covariance.
The primary distinction lies in the upper-level loss function for the parameters, detailed as follows.
\begin{enumerate}
\item \emph{Approximate posterior error~(APE)}~\cite{hu2017efficient}.
\begin{equation}\label{eq:APE}
    \mathcal{L}^{\text{APE}}(\theta) = 1/N\sum\nolimits_{k=1}^N \operatorname{tr}(\hat{P}_k ). 
\end{equation}
\item \emph{Mean squared error of state estimate~(MSE)}~\cite{qadri2024learning}.
\begin{equation}
    \mathcal{L}^{\text{MSE}}(\theta)=1/N\sum\nolimits_{k=1}^N \| \hat{T}_k\boxminus T_k^{\text{gt}}\|_2^2,
\end{equation}
where $T_k^{\text{gt}}\in SE(3)$ denote the ground truth poses, and $\hat{T}_k \in SE(3)$ is the estimation pose from the filter. 

\item \emph{Average innovartion~(Innov.)}~\cite{li2019unsupervised}. 
\begin{equation}
    \mathcal{L}^{\text{innov}}(\theta) = 1/N \sum\nolimits_{k=1}^N \|h(\xi_k)\boxminus  y_k \|_2^2.
\end{equation}
\end{enumerate}
The implementations of gradient descent optimization in~\cite{qadri2024learning,li2019unsupervised} follow the original papers, which rely on numerical gradients. The approach of~\cite{hu2017efficient} uses Bayesian optimization for general hyperparameters, whereas we focus on a differentiable noise parameter and apply gradient descent.

\begin{table*}[ht]
\label{Tab:1}
\centering
\caption{Comparison of different optimization methods across datasets.}
\renewcommand{\arraystretch}{1.2}
\setlength{\tabcolsep}{4pt}
\begin{tabular}{|l|c|c|c|c|c|c|c|c|c|}
\hline
\textbf{Dataset} & \textbf{Sensor} & \textbf{Real Noise} & \textbf{Init. Value} & $\boldsymbol{\ell^o}$ only & \textbf{APE} & \textbf{Innov.} & \textbf{MSE} & \textbf{Ours (loop-)} & \textbf{Ours (loop+)} \\
\hline
\multirow{12}{*}{Loop-to-Open}
& \multirow{3}{*}{GPS} 
& $4.00\times 10^{0}$ & $1.00\times 10^{0}$ & $3.76\times 10^{0}$ & $8.21\times 10^{-4}$ & $8.17\times 10^{1}$ & $5.92\times 10^{1}$ & $4.27\times 10^{0}$ & $4.52\times 10^{0}$  \\
&   & $9.00\times 10^{0}$ & $1.00\times 10^{0}$ & $8.04\times 10^{0}$ & $8.08\times 10^{-4}$ & $1.00\times 10^{2}$ & $1.00\times 10^{2}$ & $8.99\times 10^{0}$ & $9.78\times 10^{0}$  \\
&   & $1.00\times 10^{0}$ & $1.00\times 10^{0}$ & $8.90\times 10^{-1}$ & $1.08\times 10^{-5}$ & $1.76\times 10^{1}$ & $6.59\times 10^{0}$ & $8.80\times 10^{-1}$ & $8.70\times 10^{-1}$ \\ \cline{2-10}

& \multirow{2}{*}{IMU} 
& $1.00\times 10^{-6}$ & $1.00\times 10^{-1}$ & $5.48\times 10^{-4}$ & $1.00\times 10^{-6}$ & $1.00\times 10^{-6}$ & $1.00\times 10^{-6}$ & $1.00\times 10^{-6}$ & $1.00\times 10^{-6}$ \\
&  & $1.00\times 10^{-4}$ & $1.00\times 10^{-1}$ & $1.06\times 10^{-1}$ & $1.12\times 10^{-2}$ & $2.23\times 10^{-3}$ & $1.34\times 10^{-3}$ & $6.67\times 10^{-3}$ & $5.36\times 10^{-3}$ \\ \cline{2-10}

& \multirow{6}{*}{VO} 
& $1.00\times 10^{-4}$ & $1.00\times 10^{1}$ & $3.91\times 10^{-4}$ & $9.98\times 10^{0}$ & $9.98\times 10^{0}$ & $9.99\times 10^{0}$ & $3.30\times 10^{-2}$ & $3.30\times 10^{-4}$ \\
&   & $2.50\times 10^{-3}$ & $1.00\times 10^{1}$ & $3.08\times 10^{-3}$ & $9.99\times 10^{0}$ & $9.99\times 10^{0}$ & $9.99\times 10^{0}$ & $3.22\times 10^{-2}$ & $3.24\times 10^{-3}$ \\
&   & $1.00\times 10^{-4}$ & $1.00\times 10^{1}$ & $3.74\times 10^{-4}$ & $9.98\times 10^{0}$ & $9.97\times 10^{0}$ & $9.98\times 10^{0}$ & $3.46\times 10^{-2}$ & $3.45\times 10^{-4}$ \\
&   & $1.00\times 10^{-6}$ & $1.00\times 10^{1}$ & $9.62\times 10^{-3}$ & $5.79\times 10^{0}$ & $6.87\times 10^{0}$ & $9.98\times 10^{0}$ & $9.31\times 10^{-3}$ & $8.45\times 10^{-3}$ \\
&   & $1.00\times 10^{-6}$ & $1.00\times 10^{1}$ & $3.84\times 10^{-3}$ & $5.42\times 10^{0}$ & $4.99\times 10^{0}$ & $5.08\times 10^{0}$ & $2.39\times 10^{-3}$ & $1.67\times 10^{-3}$ \\
&   & $2.50\times 10^{-5}$ & $1.00\times 10^{1}$ & $5.76\times 10^{-2}$ & $5.19\times 10^{0}$ & $7.63\times 10^{0}$ & $8.56\times 10^{0}$ & $5.10\times 10^{-2}$ & $4.45\times 10^{-2}$ \\ \cline{2-10}

& Test MSE & -- & $3.90\times 10^{0}$ & $4.40\times 10^{-1}$ & $3.59\times 10^{0}$ & $3.30\times 10^{-1}$ & $3.20\times 10^{-1}$ & $2.40\times 10^{-1}$ & $\mathbf{2.10\times 10^{-1}}$\\
\hline

\multirow{3}{*}{Garage Dataset}
& \multirow{2}{*}{IMU} 
& $1.00\times 10^{-6}$ & $1.00\times 10^{-2}$ & $1.00\times 10^{-6}$ & $1.03\times 10^{-2}$ & $1.09\times 10^{-2}$ & $9.98\times 10^{-4}$ & $1.00\times 10^{-6}$ & $1.00\times 10^{-6}$ \\
&  & $1.00\times 10^{-4}$ & $1.00\times 10^{-2}$ & $1.51\times 10^{-5}$ & $1.04\times 10^{-2}$ & $9.98\times 10^{-4}$ & $1.00\times 10^{-3}$ & $2.78\times 10^{-5}$ &  $1.14\times 10^{-3}$ \\ \cline{2-10}
& Test MSE & -- &  $5.81\times 10^{0}$ & $2.25\times 10^{0}$ & $3.12\times 10^{0}$ & $1.54\times 10^{0}$ & $1.49\times 10^{0}$ & $9.30\times 10^{-1}$ &$\mathbf{8.10\times 10^{-1}}$ \\
\hline
\end{tabular}
\end{table*}

\subsection{Simulation}

We first evaluate the algorithm in simulation. 
The state $X_k \in SE_2(3)$ as in~\eqref{def_SE2_3} is estimated using a left-invariant extended Kalman filter (LInEKF). 
The system kinematics is given by the IMU, 
where in LInEKF the process noise covariance is 
\(\mathrm{blkdiag}(Q_g,\bm 0,Q_a)\)~\cite{hartley2020contact}, 
with gyroscope noise \(n_g \sim \mathcal{N}(\bm 0,Q_g)\) 
and accelerometer noise \(n_a \sim \mathcal{N}(\bm 0,Q_a)\).
The unary measurements given by GPS and binary measurements given by visual odometry~(VO) are specified as
\[
y^u(X_k) = p_k + \nu^u,~ 
y^b(X_k,X_{k+1}) = \exp(\nu^b)(X_k)^{-1}X_{k+1},
\]
with Gaussian noises $\nu^u\sim \mathcal{N}(0,\Sigma^u),~\nu^b\sim \mathcal{N}(0,\Sigma^b)$.
The supervisory measurements are the relative poses provided by loop closures.
The trajectory is shown in Figure.\ref{fig:Simulation}~(a). 
The first 20 seconds are used for noise covariance estimation, termed the calibration stage, with several supervisory measurements generated.
The remaining 70 seconds are used for testing, with both stages sharing the same noise parameters.

\begin{figure}[H]
    \centering
    \includegraphics[width=0.84\linewidth]{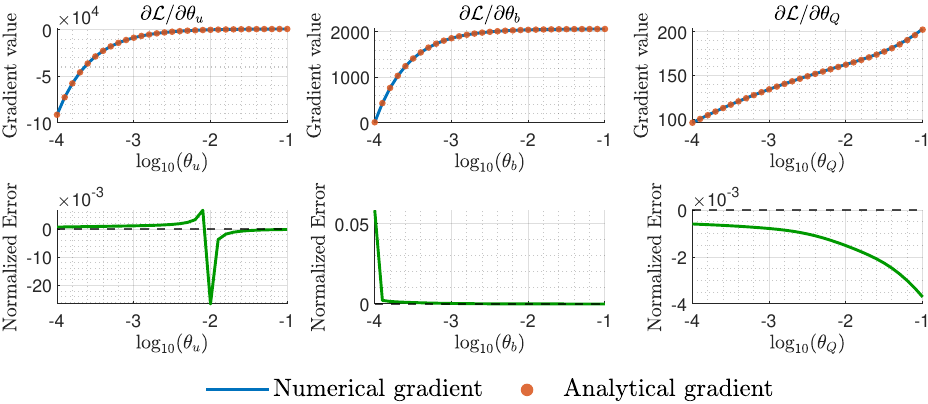}
   \caption{Comparison of the numerical and analytical gradients. The lower panel reports the normalized error $((\partial \mathcal{L}/\partial \theta_j)_{\mathrm{ana}} - (\partial \mathcal{L}/\partial \theta_j)_{\mathrm{num}}) / ((\partial \mathcal{L}/\partial \theta_j)_{\mathrm{num}})$.}
    \label{fig:sensitity}
\end{figure}

We first verify the analytical gradient against the numerical gradient, as shown in Figure.\ref{fig:sensitity}. 
The covariance is parameterized by scalars in this example, i.e., $Q_a = Q_g = \theta_Q I$, $\Sigma^u = \theta_u I$, and $\Sigma^b = \theta_b I$. 
Despite approximations in manifold operations, the derivative remains sufficiently accurate, while analytical gradients are computed more efficiently.

Secondly, we evaluate our method against baseline approaches, using the MSE on the test stage as the evaluation metric. 
Unary and binary measurement noises are modeled as diagonal matrices of size three and six. Process noise is parameterized as $Q_a=\exp(\theta_{Q_a})I$ and $Q_g=\exp(\theta_{Q_g})I$, with $\theta_j \in [-6,2]$ ensuring positivity. To assess supervisory effects, we compare 780 loop closures from 40 keyframes (\textit{loop+}) and 190 from 20 downsampled keyframes (\textit{loop-}), with the standard deviation of the loop noise set to $10^{-3}$. True parameters and results are reported in Table.1.
Innovation and MSE assess only the states and ignore covariances, though relative reliability (e.g., smaller noise for GPS $z$-axis) can still be observed. For binary measurements, all algorithms are less sensitive to the exact noise levels.
Our method, together with the $\ell^o$ formulation, offers a likelihood-based approach that yields covariances closer to the ground truth. The comparison of $\ell^o$ with and without supervisory loops shows that supervisory measurements effectively suppress odometry drift. As illustrated in Figure.\ref{fig:Simulation}~(a), while using $ \mathcal{L}^{\text{MSE}}$ reduces trajectory error most in the calibration stage but deviates from true parameters, our method attains the lowest MSE in the long open-loop test trajectory. 

\begin{figure}[h]
    \centering
    \captionsetup[subfigure]{labelformat=empty}
    \subcaptionbox{ 
  \label{fig:ATE_level}}[0.50\linewidth]{%
        \vspace{0pt}%
        \includegraphics[width=\linewidth]{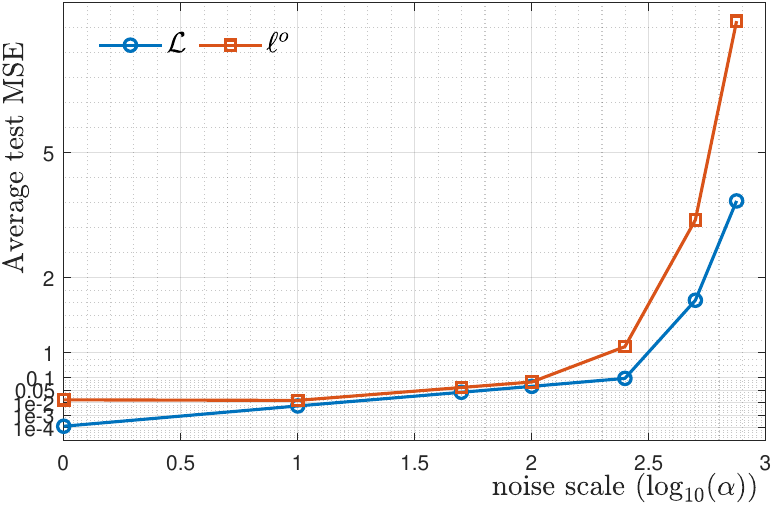}%
    }
    \subcaptionbox{ \label{tab:placeholder}}[0.40\linewidth]{%
        \centering
        \setlength{\tabcolsep}{4pt}
        \small
        \renewcommand{\arraystretch}{1.1}
        \begin{tabular}{c|c|c}
            \hline
            $\alpha$ & $\ell^o$ & $\mathcal{L}$\\
            \hline
            1   & 0.20 & \textbf{0.18}\\
            10  & \textbf{0.22} & \textbf{0.22}\\
            50  & \textbf{0.22} & 0.23\\
            100 & 0.28 & \textbf{0.26}\\
            500 & 0.83 & \textbf{0.77}\\
            750 & 1.42 & \textbf{1.39}\\
            \hline
        \end{tabular}
    }
    \vspace{-1.5em}
    \caption{Test MSE and Wasserstein error across $\alpha$.}
    \label{fig:grad_wasserstein_row}
\end{figure}

A Monte Carlo experiment is conducted to evaluate the impact of supervisory measurements by comparing $\ell^o$ and $\mathcal{L}$ as loss functions. 
Measurement noises are parameterized by a base level $\beta$ and a scaling factor $\alpha$, with $\theta_j = (\alpha \beta_j)^2$, where $\beta_j$ is sampled from $[1,2]\times10^{-4}$ for $Q$, $[2,10]\times10^{-3}$ for $\Sigma^u$, and $[1,6]\times10^{-4}$ for $\Sigma^b$. 
All algorithms are initialized identically and run for 20 iterations under the same optimizer settings. 
The average 2-Wasserstein error~\cite{khosoussi2025joint} is also reported, measuring the discrepancy between estimated and true covariances, with smaller values indicating more accurate estimation. 
As shown in Figure.\ref{fig:grad_wasserstein_row}, $\mathcal{L}$ yields consistently lower MSE than $\ell^o$ across noise levels and, under large noise, achieves smaller Wasserstein error, indicating more accurate covariance estimation. 
This demonstrates the additional benefits of incorporating supervisory measurements.

\vspace{-1em}
\subsection{Dataset Experiment}
The garage dataset~\cite{carlone2015initialization} provides real-world binary measurements and loop closures for evaluation. It features calibration segments with abundant loops and validation sections with open-loop trajectories. Lacking unary measurements, we use only IMU and binary data. Following convention, we treat the outlier-free estimation using all measurements as groundtruth.
Noisy IMU measurements are then generated from this trajectory using known noise parameters, enabling convenient evaluation of the algorithm.
As outlier rejection is not the focus of this work, we replace the false positive loop closures with measurements assigned a noise standard deviation of $0.05$. 
In the \textit{loop-} case, 136 loop pairs are used, while in the \textit{loop+} case, 234 pairs are included.

The calibration stage and ground-truth trajectory are illustrated in Figure.\ref{fig:Simulation}~(b), where large drift is observed in IMU-only or binary-only estimates.
With the initial parameters, the trajectory exhibits large drift; after calibration, accurate estimates are obtained even without absolute information from sensors such as GPS, provided that IMU and binary measurements are appropriately weighted. 
As shown in Table.1, the baseline methods $\mathcal{L}^{\text{MSE}}$ and $\mathcal{L}^{\text{innov}}$ improve state estimation but may misestimate covariances, whereas our method achieves the most effective tuning. 
Furthermore, incorporating more loop closures yields stronger supervision.

\section{Conclusion and Future Work}
This work presents a Bayesian framework for noise covariance estimation, formulated as a bilevel optimization via probability factorization. State and derivative filters run concurrently to provide estimates and analytical gradients, enabling the use of supervisory measurements for richer information with high efficiency. The method is validated on simulated and real-world datasets, and remains flexible and extensible. While the supervisory noise covariance is assumed to be known, this is not restrictive since its gradient is explicit, allowing straightforward optimization.

Several aspects of the framework warrant further exploration.
We plan to apply the algorithm to real-world SLAM systems such as~\cite{zheng2024fast}.
Besides, we will explore deployment-oriented parameterizations, for instance, differentiable neural embeddings that adapt measurement noises to context.
Moreover, the framework’s flexibility also enables tuning of broader hyperparameters, which remains to be investigated.




\bibliographystyle{unsrt}
\bibliography{reference}

\end{document}